%% file: main.tex
\documentclass[10pt,twocolumn,letterpaper]{article}

\usepackage{iccv}
\usepackage{times}
\usepackage{epsfig}
\usepackage{graphicx}
\usepackage{amsmath}
\usepackage{amssymb}
\usepackage{bm}
\usepackage{color, colortbl}
\usepackage{multirow}
\usepackage{subfig}

\graphicspath{{./figures/}}

\usepackage[ruled,linesnumbered]{algorithm2e}

\usepackage[font=small,skip=0pt,labelsep=period]{caption}

\makeatletter
\newcommand{\thickhline}{%
    \noalign {\ifnum 0=`}\fi \hrule height 0.5pt
    \futurelet \reserved@a \@xhline
}
\makeatother
\definecolor{LightGray}{gray}{0.9}
\definecolor{MyGray}{gray}{0.5}
\newcommand{\pub}[1]{\color{MyGray}{\tiny{[{#1}]}}}

\usepackage[pagebackref=true,breaklinks=true,letterpaper=true,colorlinks,bookmarks=false]{hyperref}

\iccvfinalcopy 

\def\iccvPaperID{3115} 

\ificcvfinal\fi

\begin{document}

\title{Super-Resolving Cross-Domain Face Miniatures by\\ Peeking at One-Shot Exemplar}

\author{%
Peike~Li$^{1,3}$\thanks{Part of this work is done as an intern at Baidu Research},~Xin~Yu$^{1}$,~Yi~Yang$^{2}$ \\
$^{1}$ReLER, Centre for Artificial Intelligence, University of Technology Sydney\\
$^{2}$CCAI, College of Computer Science and Technology, Zhejiang University\\
$^{3}$Baidu Research\\
}


\maketitle
\ificcvfinal\thispagestyle{empty}\fi

\begin{abstract}
\vspace{-1em}
Conventional face super-resolution methods usually assume testing low-resolution (LR) images lie in the same domain as the training ones.
Due to different lighting conditions and imaging hardware, domain gaps between training and testing images inevitably occur in many real-world scenarios. 
Neglecting those domain gaps would lead to inferior face super-resolution (FSR) performance.
However, how to transfer a trained FSR model to a target domain efficiently and effectively has not been investigated.
To tackle this problem, we develop a Domain-Aware Pyramid-based Face Super-Resolution network, named DAP-FSR network. 
Our DAP-FSR makes the first attempt to super-resolve LR faces from a target domain by exploiting only a pair of high-resolution (HR) and LR exemplar in the target domain. 
To be specific, our DAP-FSR firstly employs its encoder to extract the multi-scale latent representations of the input LR face. 
Considering only one target domain example is available, we propose to augment the target domain data by mixing the latent representations of the target domain face and source domain ones, and then feed the mixed representations to the decoder of our DAP-FSR. 
The decoder will generate new face images resembling the target domain image style. 
The generated HR faces in turn are used to optimize our decoder to reduce the domain gap.
By iteratively updating the latent representations and our decoder, our DAP-FSR will be adapted to the target domain, thus achieving authentic and high-quality upsampled HR faces.
Extensive experiments on three benchmarks validate the effectiveness and superior performance of our DAP-FSR compared to the state-of-the-art methods.
\end{abstract}


\section{Introduction}
\input{section/1-introduction}

\section{Related Work}
\input{section/2-related-work}

\section{Task Definition: One-shot based FSR}
\input{section/3-task-definition}

\section{Proposed Method}
\input{section/4-method}

\section{Experiments}
\input{section/5-experiment}

\section{Conclusion}
In this paper, we addressed a more challenging and practical face super-resolution task, where a domain gap between the training and testing data exists. 
To tackle this problem, we proposed a new Domain-Aware Pyramid-based Face Super-Resolution network (DAP-FSR) that is able to super-resolve unaligned low-resolution ones from a target domain effectively by leveraging only one target domain exemplar.
Our approach bridges the domain gap by fully exploiting the given exemplar from the target domain as well as our designed soft mixing strategy which significantly enlarges the number of the training samples. 
Extensive experiments demonstrate our method is able to super-resolve cross-domain LR faces and outperforms the state-of-the-art methods significantly. 
We hope that our work will also motivate future research on the low-shot FSR task.
\clearpage
{\small
\bibliographystyle{ieee_fullname}
\bibliography{egbib}
}


\end{document}


\title{ Supplementary Materials: \\Super-Resolving Cross-Domain Face Miniatures by\\ Peeking at One-Shot Exemplar}

\author{Peike Li\\
AAII, Univerity of Technology Sydney\\
{\tt\small peike.li@student.uts.edu.au}
\and
Xin Yu\\
AAII, Univerity of Technology Sydney\\
{\tt\small xin.yu@uts.edu.au}
\and
Yi Yang\\
AAII, Univerity of Technology Sydney\\
{\tt\small yi.yang@uts.edu.au}
}

\maketitle

\input{latex/section/appendix}


%% file: section/1-introduction.tex
Face Super-Resolution (FSR), also known as face hallucination, aims at reconstructing high-resolution (HR) face images from input low-resolution (LR) ones. FSR provides critical information for the downstream computer vision and machine learning tasks, such as face detection~\cite{Bai_2018_CVPR}, recognition~\cite{liu2017sphereface} and photo-editing~\cite{Jo_2019_ICCV,wang2020single,li2021write}.
Thanks to the advance of generative adversarial networks~\cite{goodfellow2014generative}, FSR has achieved great success in recent years~\cite{yu2016ultra,yu2017hallucinating,yu2017face,yu2018face,yu2018imagining,yu2018super,yu2019can,bulat2018super,chen2018fsrnet,zhang2020copy}. 

\begin{figure}[t]
    \centering
    \includegraphics[width=\linewidth]{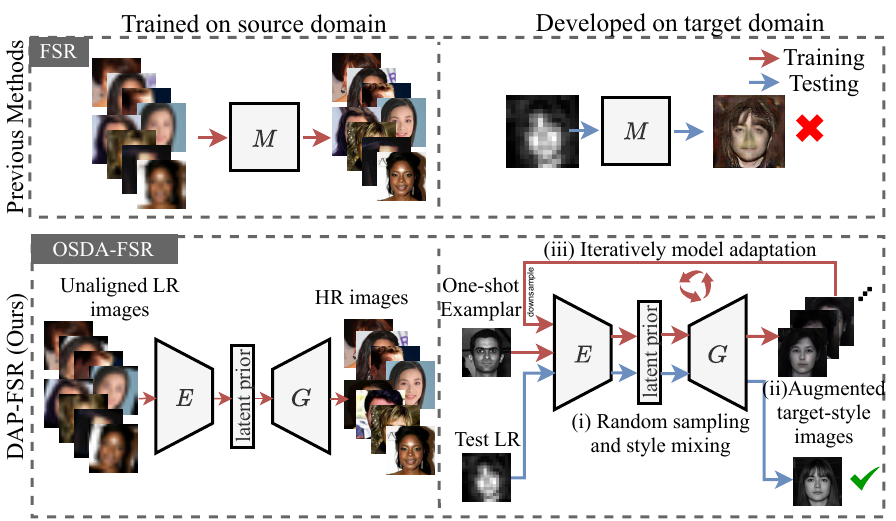}
    \vspace{2mm}
    \caption{Conventional FSR methods achieve good performance on the source dataset, but are prone to fail on the target dataset due to the domain gap. Our proposed method effectively adapts the model by leveraging only one-shot example.}
    \label{fig:motivation}
\end{figure}

Previous FSR methods usually presume training and testing LR faces are captured from the same domain. When testing LR faces resemble the training ones, previous works achieve authentic upsampled HR faces. 
However, in practice, the domain gap between testing images and training ones is inevitable due to different imaging equipment, illumination conditions, \textit{etc}.
As shown in the upper right of Figure~\ref{fig:motivation}, previous state-of-the-art FSR methods fail to upsample HR authentically due to the large domain gap between the target domain (testing) and source domain (training).
Considering FSR models would be deployed in different scenarios, it is very inefficient to re-train every deployed FSR model by collecting large-scale data from the corresponding target domain. 
Therefore, only using a few samples, ideally one example, to efficiently update an FSR model is highly desirable.

In this paper, we aim to super-resolve LR faces that exhibit an obvious domain gap by only leveraging one-shot exemplar from the target domain. We name this task as One-Shot Domain Adaption for Face Super-Resolution (OSDA-FSR). 
Different from conventional FSR methods~\cite{yu2016ultra,chen2018fsrnet,menon2020pulse}, two challenges are naturally raised:
(i) how to design a FSR network architecture that is intrinsically suitable for efficient adaptation; and (ii) how to explore one example to bridge the domain gap since simply fine-tuning an FSR network with one example is ineffective.

To address these challenges, we present a novel Domain-Aware Pyramid based Face Super-Resolution network, namely DAP-FSR network.
Our DAP-FSR contains two parts: a domain-aware pyramid encoder and an upsampling decoder.
Our DAP-FSR encoder is designed to extract the latent representations by leveraging the multi-scale features from the input LR faces.
Considering LR faces may be unaligned, we propose an Instance Spatial Transformer Networks (ISTN) to align LR faces inspired by~\cite{jaderberg2015spatial}. 
In this way, we facilitate the latent representation learning and face upsampling processes by aligning the LR faces into the canonical view. 
Motivated by the powerful architecture of StyleGAN~\cite{karras2019style, karras2020analyzing}, an image generation network, we construct our upsampling decoder.
Once we obtain the latent representations, we feed those representations to our DAP-FSR decoder to hallucinate high-quality HR face images.

To tackle the problem of super-resolving LR faces in a new domain without the need for tremendous data collection, we propose a Domain-Aware latent Mixing and Model Adaptation algorithm (DAMMA). In a nutshell, our DAMMA algorithm is able to adapt the model trained on the source domain to the target domain by exploring only the one-shot example. 
As illustrated in Figure~\ref{fig:motivation}, when a target domain example is given, DAP-FSR network first extracts its latent representations. 
Then, supervised by the given one-shot example, we learn a soft mixture weight to mix the target latent representations with random-sampled source latent ones.
In this fashion, the newly generated faces will resemble the target domain faces and we significantly augment the target-style data.
By constrained fine-tuning our decoder with the augmented images, our network is gradually adapted from the source domain to the target domain. 
After iteratively updating the soft mixing weight and adapting our decoder, our DAP-FSR attains authentic target domain HR faces. 


Our main contributions are summarized as follows,
\begin{itemize}
\vspace{-0.5em}
\item We propose a novel domain-aware pyramid-based face super-resolution network, named DAP-FSR network, to efficiently upsample cross-domain LR face images by peeking at one-shot target domain example.  
\vspace{-0.5em}
\item We present a simple yet effective domain-aware latent mixing and model adaptation algorithm (DAMMA) to adapt our DAP-FSR to the target domain. 
Our DAMMA generates target-style alike faces to adapt the upsampling decoder in DAP-FSR by fully exploiting the one-shot example.  
\vspace{-0.5em}
\item To the best of our knowledge, our method is the first attempt to super-resolve cross-domain LR face images, making our method more practical.
\vspace{-0.5em}
\item Our proposed DAP-FSR can be adapted to a target domain effectively and is also robust to unaligned LR faces. Experiments on three constructed cross-domain face super-resolution benchmarks validate the superior performance of our proposed approach compared to the state-of-the-art methods.
\end{itemize}

%% file: section/2-related-work.tex
\noindent\textbf{Face super-resolution}.
Face super-resolution (FSR), also known as face hallucination, aims at establishing the intensity relationships between input LR and output HR face images from the same domain. 

Traditional holistic appearance-based methods firstly leverage a parameterized model to represent faces and then construct the mappings between LR and HR faces. 
Some representative models super-resolve HR faces from LR ones by adopting global linear mapping~\cite{wang2005hallucinating,liu2007face}, or optimal transport~\cite{kolouri2015transport}.
However, they require input LR images to be aligned to a canonical pose and HR faces in the database to share similar facial expressions.
Later on, part-based approaches have been proposed to relax the strict requirements in holistic appearance-based methods. 
Part-based face hallucination algorithms~\cite{ma2010hallucinating,tappen2012bayesian,yang2018hallucinating} firstly extract local facial regions and then upsample them separately.

Taking advantage of the powerful feature representation of deep neural networks, deep learning based face super-resolution methods~\cite{zhu2016deep,yu2016ultra,yu2017face,yu2019semantic,zhang2021recursive,zhu2016deep,huang2017wavelet,chen2018fsrnet,menon2020pulse,zhang2021face} have been proposed and achieved promising results. 
Several methods exploit prior knowledge, such as facial attributes~\cite{yu2018super}, parsing maps~\cite{chen2018fsrnet}, facial landmarks~\cite{yu2018face,bulat2018learn,Bulat_2018_CVPR} and identity~\cite{zhang2018super,shiri2019identity}, to advance the upsampling performance.
However, when LR faces are captured from another domain, such as different imaging conditions, existing methods may fail to super-resolve them photo-realistically. Moreover, when the new domain data is not abundantly available, it would be difficult to retrain FSR networks with such a limited number of samples.
In this paper, we make the first attempt to address this challenging scenario in a data efficient manner.    

\noindent\textbf{One-shot Domain Adaptation}
To overcome the need of large-scale training data and improve the adaption ability of models on new domains, many works have been extensively proposed~\cite{finn2017model,liu2019few,motiian2017few,bousmalis2017unsupervised,benaim2018one,wang2021audio2head,liu2019learning,feng2019attract,feng2019dual,feng2020cascaded}.
Early one/few-shot based classification tasks~\cite{fei2006one} construct generative models from shared appearance priors across classes for classification. Recently, a new stream of works focuses on using meta-learning to quickly adapt models to novel tasks~\cite{finn2017model,nichol2018first,ravi2016optimization}. 
However, these one/few-shot methods are mainly applied to different classification tasks without taking domain gaps between image-pairs into account.

Pix2Pix~\cite{isola2017image} and CycleGAN~\cite{zhu2017unpaired} have been proposed as image-to-image translation networks. However, due to the scarcity of samples in the target domain, these methods might not be suitable for transferring from the source domain to the target one with few samples.
To mitigate the data hungry problem of deep neural networks, several works employ shared~\cite{liu2019few} or partially shared~\cite{lee2018diverse} latent space assumption to conduct image-to-image translation tasks, such as style transfer~\cite{huang2017arbitrary,liu2019few} and face generation~\cite{yang2020one}.
Since these methods only address the domain gap without learning the mapping between LR and HR images, they are not suitable for face hallucination.

%% file: section/3-task-definition.tex
\begin{figure*}[t]
    \centering
    \includegraphics[width=0.85\textwidth]{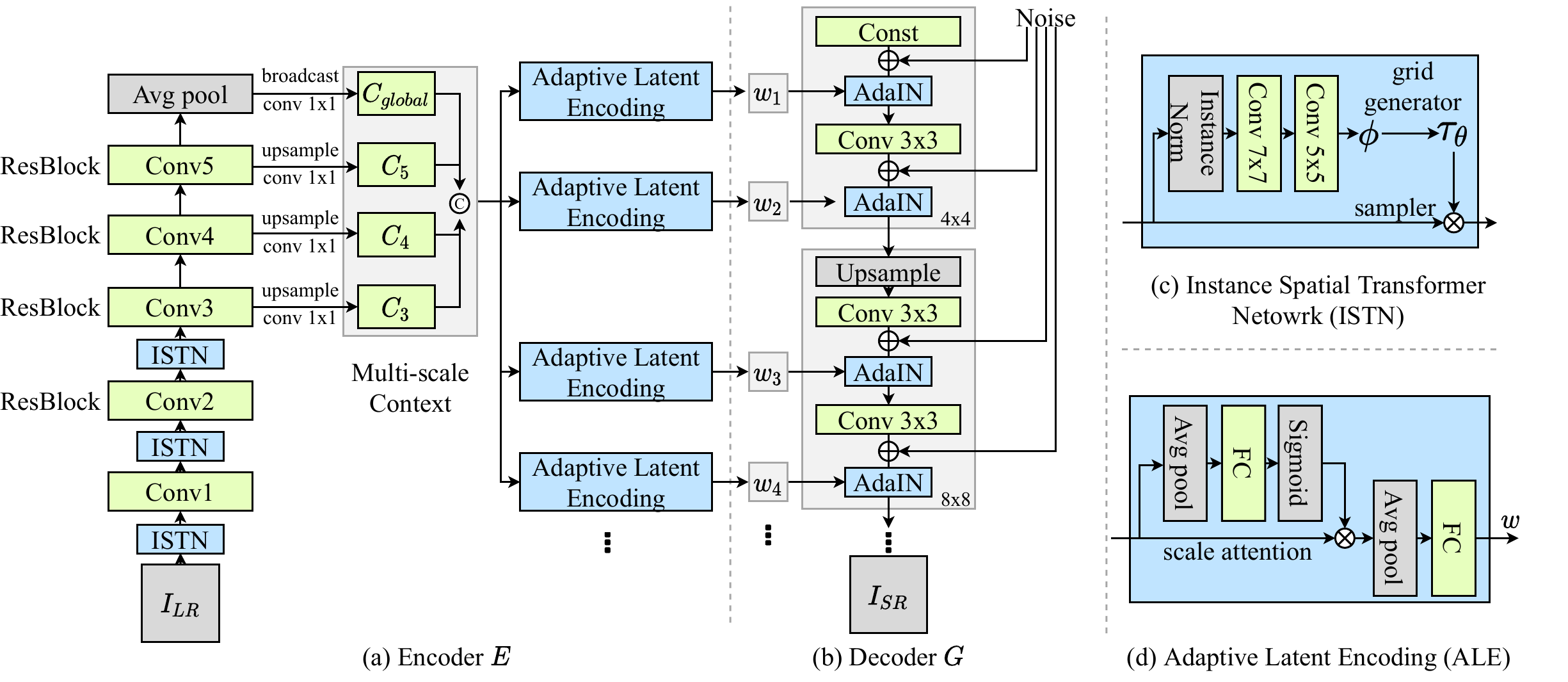}
    \vspace{2mm}
    \caption{Illustration of our DAP-FSR architecture. (a) The encoder network. Feature maps from different spatial resolution are up-sampled and concatenated as the multi-scale pyramid context. Each Adaptive Latent Encoding (ALE) module dynamically attends the multi-scale context to generate the latent representation $\bm{w}_i$. (b) The decoder network, where the HR images are generated based on the latent representations. (c) The Instance Spatial Transformer Network (ISTN) learns the style-invariant affine transformation matrix to adjust the unaligned LR images. (d) The detailed Adaptive Latent Encoding module, where the channel-wise feature attention is learned to adaptively capture the multi-scale information of the input images. }
    \label{fig:network}
\end{figure*}

Conventional Face Super-Resolution (FSR) methods aim to learn a face super-resolution model $M$ that generates a high-resolution super-resolved face image $I_{SR}\in\mathbb{R}^{H \times W}$ from a low-resolution one $I_{LR}\in\mathbb{R}^{h \times w}$, as follows:
\begin{equation}
    I_{SR}=M\left(I_{LR}\right).
\end{equation}
The goal of the FSR task is to make the reconstructed image $I_{SR}$ best recover its corresponding high-resolution version $I_{HR}$. In the conventional face hallucination setting~\cite{liu2007face,ma2010hallucinating,yu2016ultra}, an FSR model $M$ is trained and evaluated on the $\{(I_{LR}, I_{HR}) \}$ pairs from the same source domain.
However, as illustrated in Figure~\ref{fig:motivation}, when LR images come from another target domain, a pre-trained model $M$ might fail to generalize well to the new domain data and the quality of super-resolved HR images will degrade severely. 
 
Inspired by previous domain adaptation works~\cite{park2019arbitrary}, we formulate our task as One-Shot Domain Adaptation for Face Super-Resolution (OSDA-FSR). In general, OSDA-FSR can be divided into two stages, \ie, a procurement stage and a deployment stage, based on the real-world application scenario.
In the procurement stage, an FSR model is trained on the large-scale source dataset with $N_s$ HR and LR image pairs, denoted as $\mathcal{D}_s=\{(I_{LR}^s, I_{HR}^s)_i\}_{i=1}^{N_s}$. 
In this stage, image reconstruction objectives will be employed to optimize the model parameters. 
However, in the deployment stage, a trained model might encounter an unknown data distribution shift in a target domain. In this case, a deep model may fail to super-resolve LR faces in a target domain without knowing any information about the new domain.

Although collecting data and re-training a network can solve this issue, it might be inefficient and time-consuming when deploying deep models in many different real-world scenarios. 
Therefore, we aim to use only a few examples, \eg, $K$ LR-HR pairs $\mathcal{D}_t=\{(I_{LR}^t, I_{HR}^t)_i\}_{i=1}^{K}$, to effectively adapt the pre-trained model $M$. Without the loss of generality, we focus on the most challenging case where $K=1$. In other words, we will exploit the one-shot exemplar to minimize the domain gap and then hallucinate the target domain LR faces.

%% file: section/4-method.tex
\noindent\textbf{Overview.} 
The general goal of OSDA-FSR task is to transfer the model from the trained source domain to the target domain by fully exploiting the given one-shot example. 
To achieve this goal, the key idea of our approach is to adapt the model towards the target domain by enriching the target-style samples beyond the solely given one-shot exemplar.
We present a Domain-Aware Pyramid-based Face Super-Resolution (DAP-FSR) network to super-resolve input LR images to output HR images, as shown in Figure~\ref{fig:network}. 
Our DAP-FSR firstly obtains the semantic latent representations from an unaligned LR face image by the encoder network and then generates the high-quality HR images from these latent representations by the upsampling decoder network. 

Given an LR image in the target domain, our DAP-FSR network first extracts the latent representations. However, due to the existing large domain gap, the latent representations of target domain LR images may not lie on the manifold of the source domain ones, thus causing inferior upsampled results. 
To address this problem, we propose to project the latent representations of the target one-shot exemplar to the closest one in the source domain.
We then synthesize random images sharing similar styles with the target domain by mixing randomly sampled source and the extracted target domain latent representations.
These generated samples will be in turn used to optimize our upsampling network. In this fashion, the latent representation manifold will gradually shift to the target domain and we can super-resolve target domain LR images even with only one exemplar.

\subsection{Domain Aware Pyramid-based FSR}
\noindent\textbf{Choice of decoder and latent space.} 
Due to the advanced network architecture, StyleGAN~\cite{karras2019style, karras2020analyzing} obtains phenomenal high-resolution and photo-realistic images. 
Recent work~\cite{menon2020pulse} also demonstrates the possibility that employing a pre-trained StyleGAN, HR faces can be found from the given LR inputs.
More importantly, decouple the training of an encoder and a decoder would allow us to achieve larger upscaling factors while being less restricted by GPU memory.
Therefore, we choose the StyleGAN architecture as the upsampling decoder in our DAP-FSR.  

Former work~\cite{menon2020pulse} demonstrates that the multi-layer disentangled latent space $\mathcal{W}+$ in StyleGAN is more representative to depict an image than the normalized Gaussian distribution space $\mathcal{Z}$.
Furthermore, the layer-wise corresponding AdaIN modules in StyleGAN can also facilitate us to transfer domain-specific characteristics when we adapt our trained upsampling decoder to a target domain. Hence, to fully utilize the power of StyleGAN, we adopt the $\mathbf{w}\in\mathbb{R}^{l \times d_w}$ as our latent representations to better encode the LR images, where $l$ is the layer number and $d_w$ is the latent representation dimension. 

\vspace{0.5em}
\noindent\textbf{Latent representation learning.}
Unlike PLUSE~\cite{menon2020pulse} that optimizes a latent representation $\mathbf{w}\in\mathcal{W}+$ by minimizing the pixel-wise reconstruction loss between the downsampled version of upsampled HR image and the input image, we introduce an encoder to extract latent representations of the input LR faces. 
Doing so allows us to address unaligned LR faces and handle the domain gap by fine-tuning our upsampling decoder, while PLUSE cannot handle the domain gap and face misalignments as its decoder (\ie, pre-trained StyleGAN) is fixed and only $\mathbf{w}$ is updated during iterations.

Recall that in the StyleGAN, each latent representation controls a certain level of image details.
Hence, our encoder aims to adaptively predict latent representations from an enhanced multi-scale context feature.
Toward this goal, we develop an adaptive latent encoding (ALE) module that is able to generate latent representations for the upsampling decoder at different scales adaptively.

Here, we employ ResNet50 as our encoder to extract multi-scale feature maps at the \texttt{conv3}, \texttt{conv4}, \texttt{conv5} and average pooling layers, denoted as $C_3, C_4, C_5, C_{global}$, as shown in Figure~\ref{fig:network}. Then, each ALE generates multi-scale latent representations $\mathbf{w}_i$ for the decoder by attending the multi-scale features adaptively. Then, the latent representations are fed to our upsampling decoder for face hallucination.

\vspace{0.5em}
\noindent\textbf{Robust against unaligned LR faces.} 
Previous face hallucination methods~\cite{Liu_2017_CVPR,ma2010hallucinating,menon2020pulse} often assume LR faces are precisely aligned beforehand. However, such an assumption hardly holds in real application scenarios. 
Inspired by the works~\cite{yu2017hallucinating,zhang2020copy}, we estimate the transformation of LR images and warp them to the canonical position by the spatial transformation network (STN)~\cite{jaderberg2015spatial}.
Therefore, our network is robust against unaligned LR faces with in-plane rotations, translations and scale changes. The detailed architecture of spatial transformation layers are illustrated in Figure~\ref{fig:network}(c). 

More importantly, unlike previous FSR models~\cite{yu2017hallucinating,zhang2020copy} that use STNs, we apply an instance normalization layer to the feature maps before computing the transformation parameters in our instance spatial transformer network (ISTN) module. This allows us to obtain style-invariant feature maps. Therefore, even when target-domain LR faces are provided, our ISTN layers are still able to align them to the up-right position, potentially facilitating the following domain adaptation process.
Thus, our decoder can focus on super-resolving high-quality HR faces while preserving the latent representations from being affected by misaligned input LR faces.

\begin{algorithm}[t]
\caption{Domain-Aware Latent Mixing and Model Adaptation}\label{alg:one-shot}
\KwIn{Initialized DAP-FSR model $M=(E,G)$ trained on source dataset $\mathcal{D}_s$, one-shot exemplar $\{I^t_{LR}, I^t_{HR}\} \in \mathcal{D}_t$, initialized latent code mixing weight $\bm{\alpha}_0$, AdaIN parameter $\phi$ in $G$ ,learning rate $\xi, \eta$}
\KwOut{Adapted model $M_{\phi^*}$}
\While{do not converge}{
Generate $\mathbf{w}^t$ by manifold preserving projection as Eq.~\eqref{eq:mean_var}; \\
Sample a batch of source latent codes: $\mathbf{w}_s=\mu_{\mathbf{w}}+\sigma_{\mathbf{w}}\epsilon, \epsilon\sim\mathcal{N}(0,1)$; \\
Initialize latent code mixing weight: $\bm{\alpha} \leftarrow \bm{\alpha}_0$; \\
\For{i=1,2,3,...,n}{
Update mixing weight by Eq.~\eqref{eq:weight_loss}: $\bm{\alpha} \leftarrow \bm{\alpha}-\xi\nabla_{\bm{\alpha}}\mathcal{L}(\bm{\alpha})$; \\
Generate mixing latent codes $\mathbf{w}^m$ by Eq.~\eqref{eq:style_mixing}; \\
Update model parameters by Eq.~\eqref{eq:model-loss}: $\phi \leftarrow \phi-\eta \nabla_{\phi}\mathcal{L}(\phi)$; \\
}
}
Return final model weight $\phi$ as $\phi^*$;
\end{algorithm}

\vspace{0.5em}
\noindent\textbf{Manifold preserving encoding.}\label{sec:preseving-encoding}
Previous work~\cite{zhu2020domain} shows that it is possible to invert an arbitrary image, even not a face image, into style latent space $\mathcal{W}+$. However, such deduced latent codes are not aligned with the semantic knowledge prior learned by $G(\cdot)$ and lose the versatile image editing capability.
In our OSDA-FSR task, the situation will become even worse where a domain gap between the source and target domains exists. 

To overcome these drawbacks, we explicitly constrain the output of the encoder $E(\cdot)$ in the feature space of $G$. Particularly, instead of directly predicting the style latent codes, we predict the offset scale \textit{w.r.t.} the mean $\mu_{\mathbf{w}}$ and variance $\sigma_{\mathbf{w}}$ of the latent representations of $G$. 
To be specific, our DAP-FSR model maps the encoded representations of LR images to the latent representations $\mathbf{w}$ of the decoder, as follows:
\vspace{-0.5em}
\begin{equation}
\label{eq:mean_var}
    \mathbf{w} = \mu_{\mathbf{w}} + E(I_{LR})\sigma_{\mathbf{w}},
    \vspace{-0.5em}
\end{equation}
where $\mu_{\mathbf{w}}$ and $\sigma_{\mathbf{w}}$ are fixed during the encoder training process.
Therefore, using Eq.~\eqref{eq:mean_var}, we can explicitly constrain the latent representation output by our encoder to lie in the latent representation space $\mathcal{W}+$ of our decoder $G$.

\vspace{0.5em}
\noindent\textbf{Network optimization.}
Our encoder $E$ is trained using two losses. We employ the pixel-wise reconstruction loss $\mathcal{L}_{mse}$ to enforce reconstructed HR images to be close to their HR ground-truth $I_{HR}$, 
\begin{equation}
\mathcal{L}_{mse}=\|I_{HR}-G(\mathbf{w}))\|_{2}.
\end{equation}
In addition, we also introduce the perceptual loss to enforce the feature-wise similarity,
\begin{equation}\label{eq:percept}
\mathcal{L}_{percept}=\|F(I_{HR})-F(G(\mathbf{w}))\|_{2},
\end{equation}
where $F$ denotes the perceptual feature extractor. In our experiments, we extract features from \texttt{relu1\_1}, \texttt{relu2\_1}, \texttt{relu3\_1}, \texttt{relu4\_1} layers in VGG-19 with equal weights. 
In our final objective, we also treat the image intensity similarity and feature similarity equally, and the objective is defined as,
\begin{equation}\label{eq:model-loss}
\mathcal{L}(\theta)=\mathcal{L}_{mse}+\mathcal{L}_{percept},
\end{equation}
where $\theta$ is the trainable parameters of our network.
Note that our upsampling decoder and encoder are trained individually and thus our decoder is fixed during training our encoder.
 
\begin{figure}[t]
    \centering
    \includegraphics[width=0.9\linewidth]{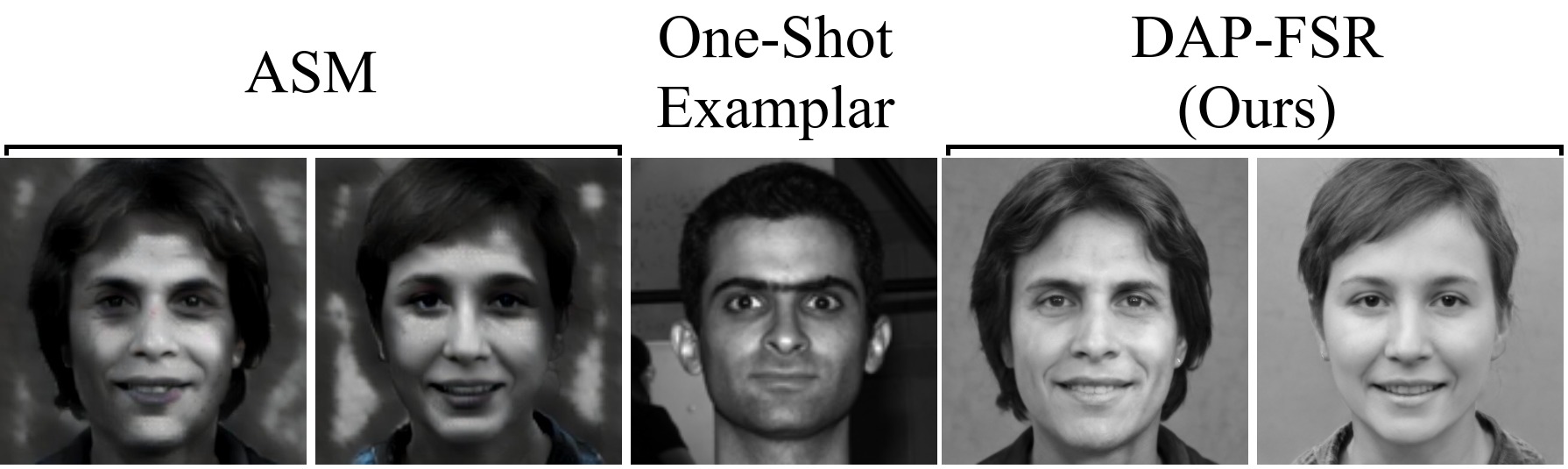}
    \vspace{2mm}
    \caption{{Compared to the style-transfer based method ASM~\cite{luo2020adversarial} (left), given only one-shot target domain exemplar  (ExtendedYaleB), our method (right) efficiently generates authentic target-style images from the source domain (CelebA). 
    }}
    \label{fig:style-transfer}
\end{figure}

\subsection{Peeking at One-Shot Exemplar}
\vspace{0.5em}
\noindent\textbf{Towards target-domain image generation.} 
Benefiting from the encoder design in our DAP-FSR network, we can encode the given one-shot target domain HR image $I_{HR}^t$ into a latent representation $\mathbf{w}^t$. However, using only one-shot exemplar does not suffice to transfer our decoder to the target domain, and will lead to an over-fitting problem. 
As explained in~\cite{karras2019style}, the latent codes of the StyleGAN control the coarse, medium, fine attributes of generated images at different style layers. 
Thus, we also regard the latent code $\mathbf{w}^t$ as an interpretable representation of a target domain face. 
Moreover, we can generate a large number of domain-specific (\ie, style-consistent) face images with $I_{HR}^t$. Specifically, for a latent code $\mathbf{w}^s$ randomly sampled from the latent representation manifold of the source domain, we mix it with $\mathbf{w}^t$ in a layer-wise manner so that a generated image $I^m$ inherits the the target domain style from $I^t$. The mixing procedure is defined as: 
\vspace{-1mm}
\begin{equation}\label{eq:style_mixing}
    \mathbf{w}_i^m = (1-\bm{\alpha}_i)\mathbf{w}_i^t + \bm{\alpha}_i\mathbf{w}_i^s,
    \vspace{-1mm}
\end{equation}
where $\bm{\alpha} \in \mathbb{R}^l$ is a layer-wise soft weight for mixing latent representations. In this manner, we effectively enlarge the number of target domain examples from the given one-shot exemplar by $G(\mathbf{w}^m)$. 
In Figure~\ref{fig:style-transfer}, compared with 
a style transfer based method (\ie, ASM~\cite{luo2020adversarial}), our method is able to generate more natural style-consistent images while preserving the identity.

\input{tables/sota-comparision}

\vspace{0.5em}
\noindent\textbf{Learning soft mixing weight.} 
When mixing the latent representations of random sampled $\mathbf{w}^s$ and the target sample $\mathbf{w}^t$, we preserve the image content information by applying a feature-wise intensity consistent loss $\mathcal{L}_{c}$ and enforce the domain information to be transferred by employing a style similarity loss $\mathcal{L}_{s}$. Here, we learn a soft weight $\bm{\alpha}$ to mix the latent codes of the source and target domain instead of manually selecting a certain layer, and the optimization process is formulated as,
\vspace{-1em}
\begin{equation}\label{eq:weight_loss}
\mathcal{L}(\bm{\alpha})=\mathcal{L}_{c}+\mathcal{L}_{s},
\end{equation}
\begin{equation}
\mathcal{L}_{c}=\|F(G(\mathbf{w}^m))- F(G(\mathbf{w}^s))\|_{2},
\end{equation}
\begin{equation}
\begin{split}
\mathcal{L}_{s}=&\|\mu(F(I^t)-\mu(F(G(\mathbf{w}^m)))\|_{2}+ \\
&\|\sigma(F(I^t))-\sigma(F(G(\mathbf{w}^m)))\|_{2}.
\end{split}
\end{equation}
where $\mu$ and $\sigma$ denote the mean and variance of the extracted features respectively, and $F$ is the same perceptual extractor in Eq.~\eqref{eq:percept}.

\vspace{0.5em}
\noindent\textbf{Model updating by constrained adaptation.} After we generate a batch of random images exhibiting the same target domain style, our next step is to adapt our model towards the target domain. The most straightforward way is to fine-tune the entire decoder $G$ directly on our generated target-domain alike samples. 
However, when the number of training examples is limited, especially in our case, fine-tuning the whole network weights often leads to over-fitting and may potentially destroy the learned knowledge prior in $G$. Instead of fine-tuning the entire decoder weights, we constrain the fine-tuning on a subset of the decoder parameters. To be specific, we only adapt the affine transform parameters in the AdaIN module. By restricting the trainable parameters, our model can be effectively adapted to the target domain while preserving the semantic knowledge, \ie, natural face structure.
The overall pipeline of our algorithm is illustrated in Algorithm~\ref{alg:one-shot}. 

\subsection{Training and Inference}
Our training process consists of two main stages, the procurement stage and development stage. In the procurement stage, we first train our decoder $G$ following the protocols of StyleGAN and then only train the encoder model $E$ on the source dataset by Eq.~\eqref{eq:model-loss} while fixing the parameters of $G$. After training, our DAP-FSR is able to super-resolve HR faces from LR faces with an upscaling factor up to $\times64$. 
In the development stage, we peek at the one-shot exemplar from the target domain and adapt our model to the target domain by employing our proposed Algorithm~\ref{alg:one-shot}.
During inference, we test our adapted model on the whole target dataset and report the super-resolution performance. Note that, we only see one-shot image from the target domain and all other testing images are \emph{never seen} during training.

%% file: tables/sota-comparision.tex
\begin{table*}[t]
\centering
\caption{Comparison with state-of-the-art methods. Results are reported on three benchmarks noted as \textbf{source $\rightarrow$ target}.
`Source only' denotes the methods only using source dataset for training, while `one-shot' denotes the methods exploring one-shot exemplar on the target dataset.
$\uparrow$ indicates that higher is better, and $\downarrow$ that lower is better.}\label{tab:sota-comparision}
\vspace{2mm}
\footnotesize{
\setlength\tabcolsep{4pt}
\renewcommand\arraystretch{1.15}
\begin{tabular}{|p{2.5pt}|l|cccc|cccc|cccc|}
\thickhline
\rowcolor{LightGray}
& & \multicolumn{4}{c|}{\textbf{CelebA $\rightarrow$ ExtendedYaleB}} & \multicolumn{4}{c|}{\textbf{CelebA $\rightarrow$ MultiPIE}}   & \multicolumn{4}{c|}{\textbf{MultiPIE $\rightarrow$ ExtendedYaleB}} \\
\rowcolor{LightGray}
& \multirow{-2}{*}{\textbf{Method}} & LPIPS $\downarrow$ & FIQ $\uparrow$ & PSNR $\uparrow$ & SSIM $\uparrow$ & LPIPS $\downarrow$ & FIQ $\uparrow$ & PSNR $\uparrow$ & SSIM $\uparrow$ & LPIPS $\downarrow$ & FIQ $\uparrow$ & PSNR $\uparrow$ & SSIM $\uparrow$  \\ 
\hline
\multirow{5}{*}{\rotatebox{90}{source only}} & Bicubic & $0.52$ & $0.31$ & $19.94$ & $0.46$ & $0.55$ & $0.27$ & $17.11$ & $0.39$ & $0.54$ & $0.31$ & $17.70$ & $0.43$ \\
& PUSLE~\cite{menon2020pulse}\pub{CVPR'20} & $0.40$ & $0.38$ & $20.18$ & $0.46$ & $0.46$ & $0.36$ & $14.63$ & $0.37$  & $0.42$ & $0.27$ & $17.02$ & $0.46$ \\
& MTDN~\cite{yu2020hallucinating}\pub{IJCV'20} & $0.39$ & $0.32$ & $17.74$ & $0.45$ & $\textbf{0.38}$ & $0.38$ & $18.00$ & $0.52$ & $0.47$ & $0.20$ & $18.67$ & $0.43$ \\
& CPGAN~\cite{zhang2020copy}\pub{CVPR'20} & $0.40$ & $0.28$ & $17.03$ & $0.47$ & $0.40$ & $0.31$ & $18.61$ & $0.52$  & $0.45$ & $0.24$ & $18.80$ & $0.44$ \\ \cline{2-14}
& DAP-FSR (Ours) & $\textbf{0.38}$ & $\textbf{0.41}$ & $\textbf{20.39}$ & $\textbf{0.49}$ & $\textbf{0.38}$ & $\textbf{0.40}$ & $\textbf{19.15}$ & $\textbf{0.54}$ & $\textbf{0.41}$ & $\textbf{0.34}$ & $\textbf{19.28}$ & $\textbf{0.46}$ \\ \hline
\multirow{4}{*}{\rotatebox{90}{one-shot}} & PULSE+ASM~\cite{luo2020adversarial}\pub{NeurIPS'20} & $0.44$ & $0.32$ & $20.47$ & $0.47$ & $0.49$ & $0.32$ & $17.87$ & $0.41$ & $0.44$ & $0.23$ & $17.89$ & $0.43$ \\
& MTDN+ASM  & $0.42$ & $0.27$ & $19.01$ & $0.48$ & $0.44$ & $0.33$ & $19.38$ & $0.53$ & $0.52$ & $0.25$ & $19.11$ & $0.47$ \\
& CPGAN+ASM & $0.49$ & $0.26$ & $18.42$ & $0.42$  & $0.49$ & $0.29$ & $19.29$ & $0.55$ & $0.51$ & $0.23$ & $19.19$ & $0.49$ \\ \cline{2-14}
& DAP-FSR (Ours) & $\textbf{0.36}$ & $\textbf{0.46}$ & $\textbf{22.32}$ & $\textbf{0.55}$ & $\textbf{0.36}$ & $\textbf{0.44}$ & $\textbf{21.00}$ & $\textbf{0.61}$ & $\textbf{0.39}$ & $\textbf{0.40}$ & $\textbf{20.43}$ & $\textbf{0.51}$ \\
\thickhline
\end{tabular}
}
\vspace{-4mm}
\end{table*}

%% file: section/5-experiment.tex
In this section, we conduct extensive experiments to evaluate our DAP-FSR framework. Since we focus on the OSDA-FSR task, we mainly compare with the state-of-the-art in this scenario. 

\subsection{Datasets and Evaluation Protocols}
\noindent\textbf{Benchmarks.} 
Current FSR benchmarks conduct training and testing within the same domain, and do not support the setting of the cross-domain OSDA-FSR task. 
Therefore, We propose three benchmarks to evaluate the performance of our DAP-FSR, \ie, CelebA~\cite{liu2015faceattributes} $\rightarrow$ Multi-PIE~\cite{Multipie}, CelebA $\rightarrow$ ExtendedYaleB~\cite{yaleb}, and Multi-PIE $\rightarrow$ ExtendedYaleB.
In particular, CelebA dataset contains large-scale in-the-wild face images, Multi-PIE and ExtendedYaleB datasets comprise indoor face images captured in different poses and illumination conditions. 
We select 10 different illumination and pose condition data splits in Multi-PIE and ExtendedYaleB, respectively.
The adaptation performance is evaluated with a given exemplar in each split and then the final reported performance is averaged over all the splits. 

\noindent\textbf{Evaluation metrics.} We report the quantitative results using the average Peak Single-to-Noise Ratio (PSNR), Structural SIMilarity scores (SSIM) following the common FSR practice~\cite{yu2017hallucinating,zhang2020copy}. 
Furthermore, we also employ the Learned Perceptual Image Patch Similarity (LPIPS)~\cite{zhang2018perceptual} and Face Image Quality (FIQ)~\cite{hernandez2019faceqnet} to evaluate the quality and authenticity of super-resolved faces. 
The PSNR, SSIM, LPIPS metrics are calculated between the reconstructed HR images $I_{SR}$ and the ground-truth HR images $I_{HR}$. The FIQ is a non-reference metric for face quality assessment, which is calculated only on $I_{SR}$.

\begin{figure*}[thb]
    \centering
    \includegraphics[width=0.90\textwidth]{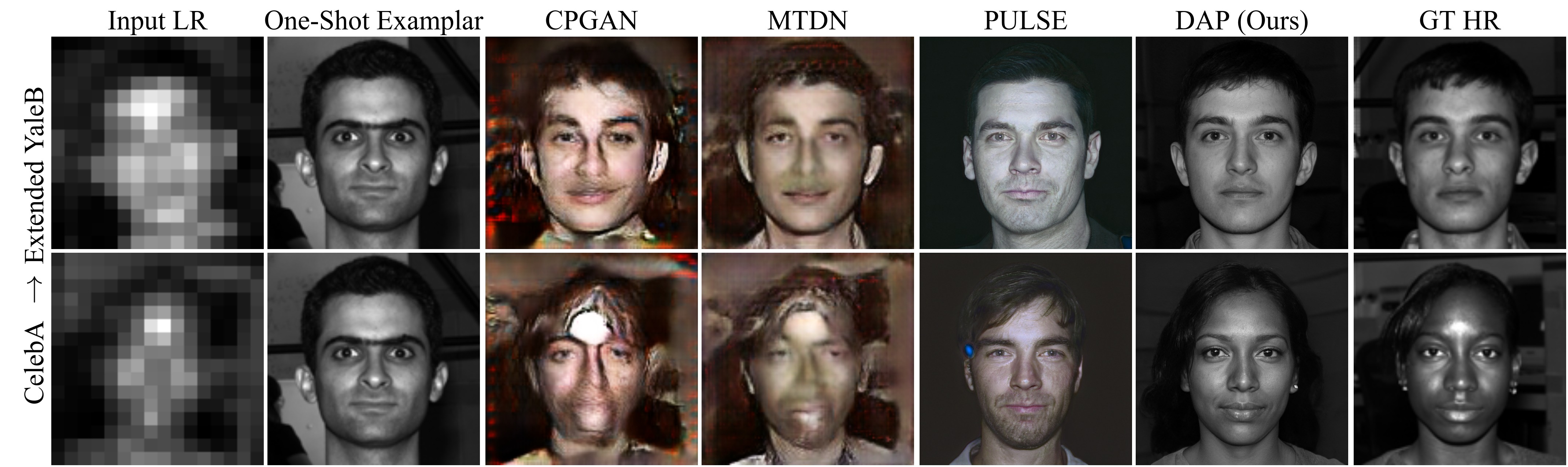}
    \includegraphics[width=0.90\textwidth]{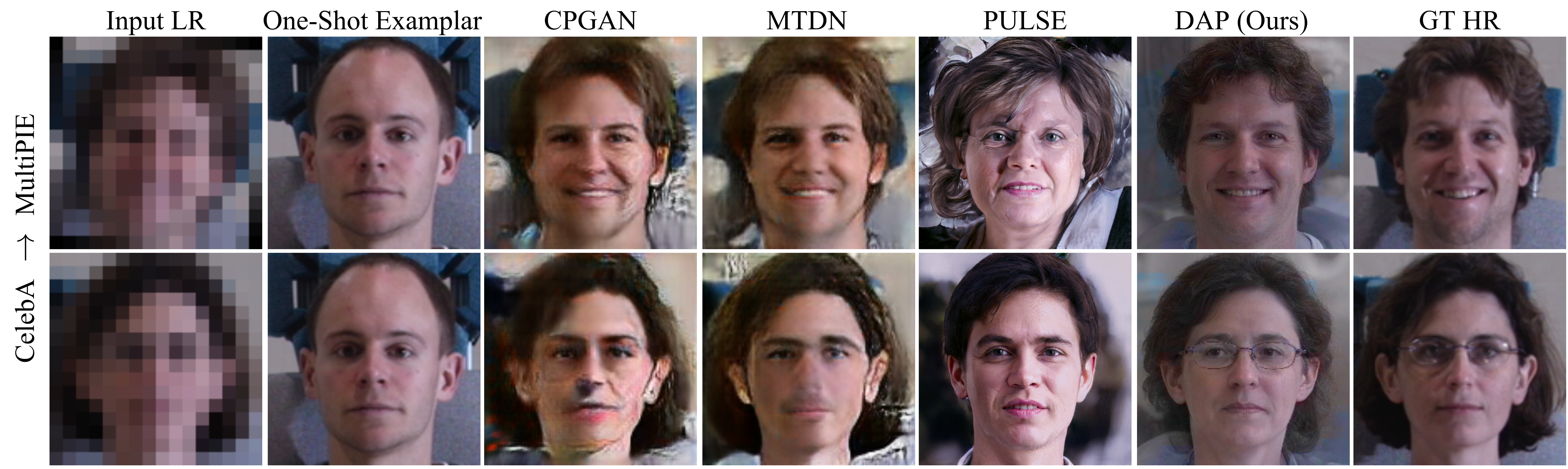}
    \includegraphics[width=0.90\textwidth]{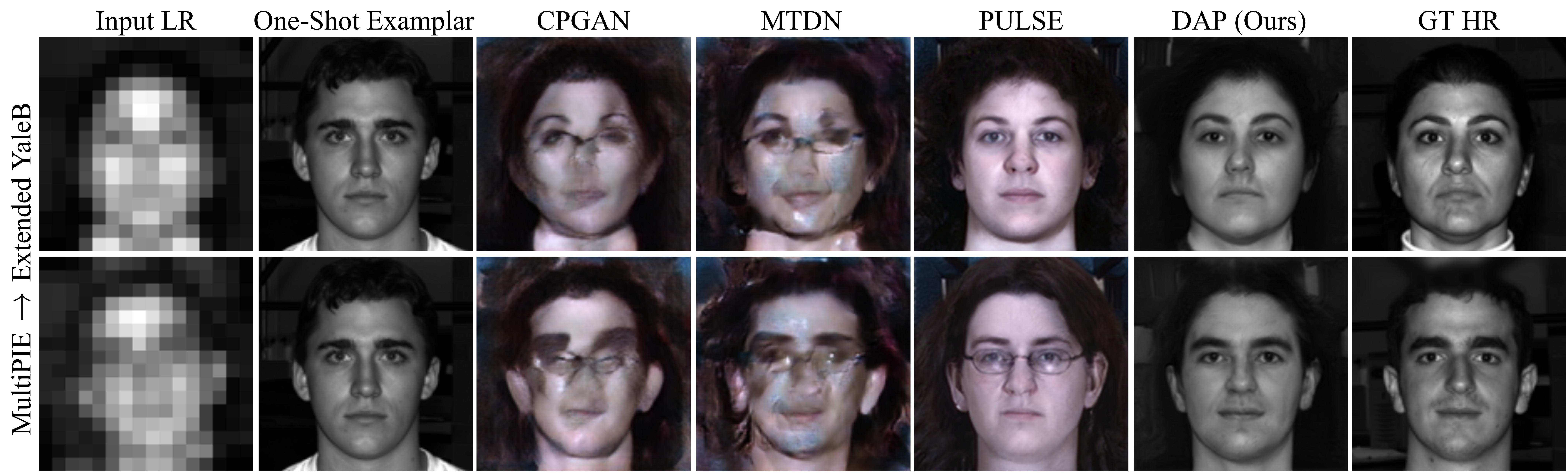}
    \vspace{2mm}
    \caption{Comparisons with state-of-the-art methods on CelebA$\rightarrow$ExtendedYaleB, CelebA$\rightarrow$MultiPIE and MultiPIE$\rightarrow$ExtendedYaleB benchmarks under the OSDA-FSR setting. Our method achieves high-quality, style-consistent HR faces and is also robust against unaligned LR inputs.}\label{fig:main-results}
    \vspace{-6mm}
\end{figure*}

\subsection{Implementation Details} 
In our experiments, we crop the aligned faces and resize them to $128\times128$ pixels to achieve ground-truth HR images. In real-world applications, we do not assume that the input LR faces are perfectly aligned. 
Following \cite{yu2020hallucinating}, we apply affine transformations, including rotations, translations and scaling, to HR faces and then downsample them to $16\times16$ pixels as our LR face images. 
We use the author-provided codes of PULSE~\cite{menon2020pulse}, MTDN~\cite{yu2020hallucinating} and CPGAN~\cite{zhang2020copy}. For comparison fairness, we adopt the same training protocols for all the methods. 
To alleviate the influence of the selected one-shot exemplar, we run the proposed method for ten times with different randomly selected one-shot exemplars in each task and report the averaged results. 

\subsection{Comparisons with the State-of-the-Art}
\noindent\textbf{Qualitative comparisons.} 
We first conduct qualitative comparisons with the state-of-the-art methods on three OSDA-FSR benchmarks in Figure~\ref{fig:main-results}.

CPGAN~\cite{zhang2020copy} and MTDN~\cite{yu2020hallucinating} can super-resolve LR images well and deal with unaligned LR input faces successfully in the source domain. However, these methods do not take the domain gap into account, and lack an efficient mechanism to address LR images from a new domain. Therefore, their final reconstructed HR images from target domain LR faces suffer from severe artifacts.
Although collecting a large number of target domain data and then re-training the networks can solve the above issue, doing so is time-consuming and does not provide a data-efficient solution to OSDA-FSR.

PULSE~\cite{menon2020pulse} traverses the high-resolution face image manifold and searches images whose downsampled versions are close to the given LR images. Although realistic images are achieved, this method requires input LR images to be perfectly pre-aligned. When LR images are unaligned, the reconstructed HR images are enforced to match the intensities of LR faces. This will lead to severe changes of face identities, as seen in Figure~\ref{fig:main-results}. Moreover, PULSE does not consider the domain gap. Due to the data distribution shift between the source and target domains, PULSE fails to super-resolve HR faces sharing the same style as the target domain images.

In contrast, as seen in Figure~\ref{fig:main-results}, our method achieves superior performance compared to the other competing methods. Although input LR images are unaligned, our DAP-FSR still produces visually appealing HR faces which are close to their HR ground-truth. Notably, our upsampled faces also exhibit style-consistency with respect to the given one-shot target domain exemplar. This demonstrates the transfer ability of our method.
Note that our method is actually able to super-resolve LR faces with an upscaling factor up to 64$\times$, and for fair comparisons with the state-of-the-art methods, we only show HR faces in the same resolution as other methods.
To the best of our knowledge, our DAP-FSR network is \textit{the first attempt to super-resolve cross-domain LR images with only one target-domain exemplar}, and achieves superior super-resolution results. 

\begin{figure}[t]
\centering
\includegraphics[width=0.85\linewidth]{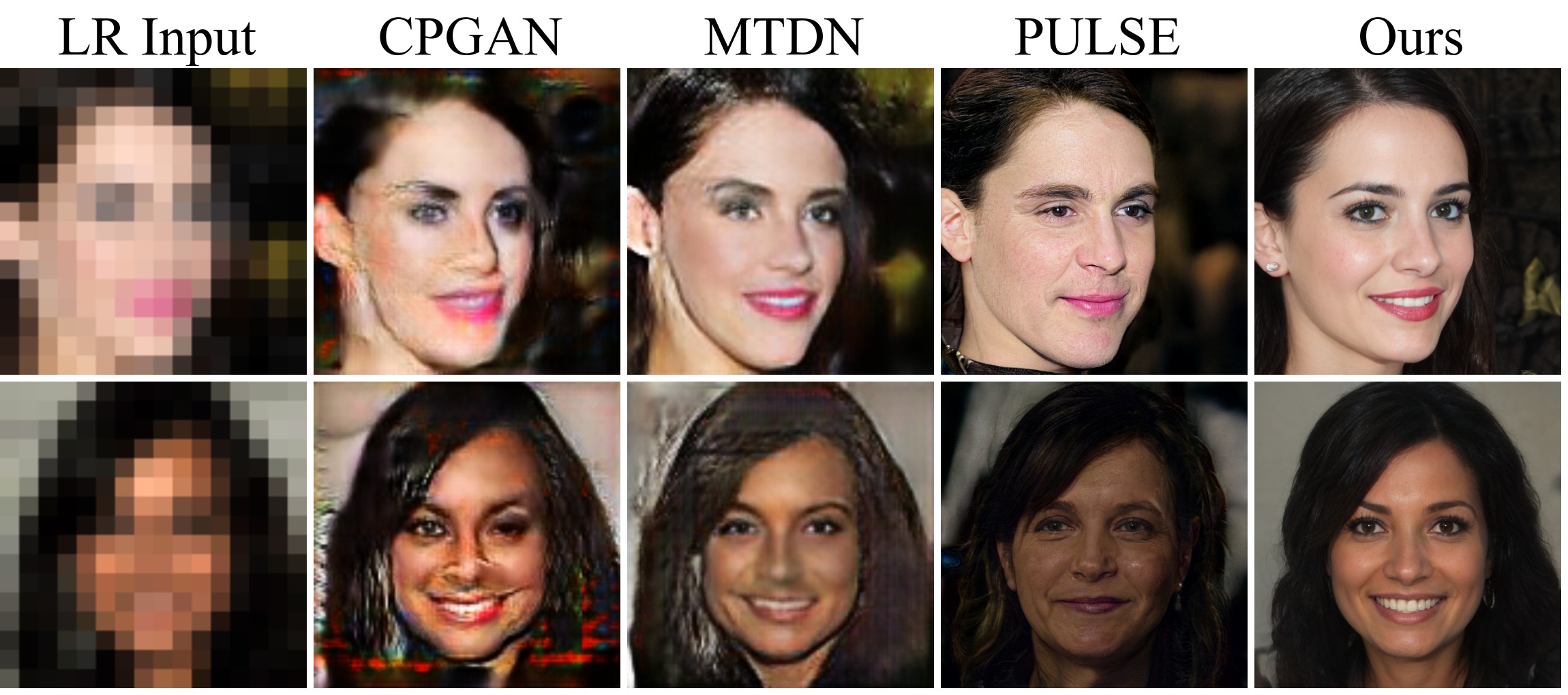}
\vspace{2mm}
\caption{Comparisons with state-of-the-art methods on tiny faces in-the-wild~\cite{bai2018finding} under real-world unconstrained conditions.}
\label{fig:tiny-face-in-the-wild}
\end{figure}

\begin{figure}[t]
\centering
\includegraphics[width=0.85\linewidth]{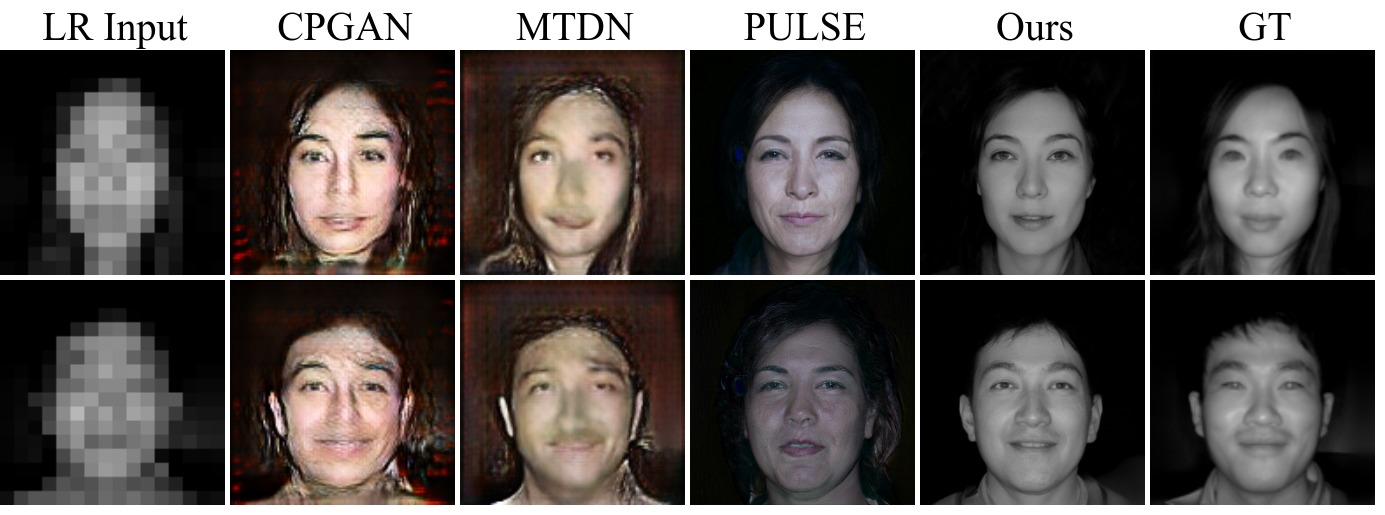}
\vspace{2mm}
\caption{Comparisons with state-of-the-art methods on near-infrared (NIR) sensor captured faces~\cite{li2013the}.}
\label{fig:nir-face}
\vspace{-4mm}
\end{figure}

To further validate the generalization ability, in Figure~\ref{fig:tiny-face-in-the-wild}, we show the FSR results of tiny faces \textit{in-the-wild}~\cite{bai2018finding} under \textit{real-world unconstrained conditions}, where the ground-truth HRs are unavailable. Here, LR faces may undergo different poses, blurs, noises, etc. All the models are trained on the CelebA source dataset and adapted to the target domain using the given one-shot HR example. 
Moreover, in Figure~\ref{fig:nir-face}, we also conduct cross-domain FSR experiments on near infrared (NIR) face images~\cite{li2013the} as a target domain.
Our DAP-FSR still outperforms the other competing methods, demonstrating the generalization ability of our method.

\noindent\textbf{Quantitative comparisons.}
As indicated in Table~\ref{tab:sota-comparision}, we report the LPIPS, FIQ, PSNR and SSIM metrics on three OSDA-FSR benchmarks, respectively.
Our proposed DAP-FSR outperforms the state-of-the-art methods significantly, especially on the perceptually-driven metrics, \ie, LPIPS and FIQ. This indicates that our super-resolved target domain HR faces not only resemble their ground-truth but also are photo-realistic. More importantly, our DAP-FSR consistently performs better than other methods on all the benchmarks. Thanks to our dedicated network design, we are able to align and upsample target domain LR faces, simultaneously. In particular, DAP-FSR reconstructs high-quality face images and outperforms the second best method PULSE on unaligned images by a margin of $+43\%$ ($0.32 \rightarrow 0.46$ ) in FIQ on the benchmark CelebA$\rightarrow$ExtendedYaleB.

To address the domain gap, a straightforward idea is fine-tuning the source-trained FSR model with the augmented target samples. Thus, we employ a style-transfer-based method ASM~\cite{luo2020adversarial} to augment new training samples from the one-shot target domain exemplar, and then fine-tune the FSR models. We name these the combination as +ASM in Table~\ref{tab:sota-comparision}. As indicated by Table~\ref{tab:sota-comparision}, applying style transfer cannot fully establish the facial detail correspondences between the source and target domains, thus leading to performance degradation.

Furthermore, benefiting from our designed one-shot adaptation algorithm, we transfer our network to the target domain effectively. Therefore, our quantitative results are better than the results of MTDN+ASM and PULSE+ASM. 
Owing to our encoder-decoder design, our method is also more efficient and effective compared to the decoder-only based method PULSE. 
After training, our DAP-FSR hallucinates LR faces in a feed-forward manner and runs $\times150$ faster than PULSE, which provides a high application potential in the real-world scenario.


\subsection{Ablation Analysis}
In our ablation analysis, 
we conduct all the experiments on the CelebA$\rightarrow$ExtendedYaleB benchmark. 

\noindent\textbf{Effectiveness of network design.} We analyze the effect of each component in our network design in Table~\ref{tab:ablations}. Compared to a straightforward approach that predicts the latent representations at the end of the backbone, our network adaptively explores the abundant multi-scale features (Config A). It is a long-standing shortcoming that CNN is sensitive to rotations. Our multiple ISTN design effectively handle this problem (Config B), thus being robust against unaligned LR images. We also illustrate that it is vital to explicitly constrain the predicted latent representations on the manifold (Config C).

\noindent\textbf{Effectiveness of one-shot domain adaptation.} Table~\ref{tab:ablations} indicates the impact of each component in Algorithm~\ref{alg:one-shot} on the OSDA-FSR performance. In our method, we effectively enrich the training samples by mixing the latent representations between the source and target domain faces (Config D). Compared to the configuration without exploring the one-shot exemplar (Config C), we observe that Config D achieves better super-resolution performance. This implies our method fully exploits the one-shot target exemplar to bridge the domain gap.

By applying the soft mixing weight (Config E), we further improve the super-resolution performance. This indicates that our soft mixing strategy is more effective than simply replacing the last three final layers of the latent representations between the source and target domain images as done in Config D.
As fine-tuning the whole decoder network may lead to over-fitting and destroy the learned face priors, we constrain the optimization space and only modify the AdaIN parameters to improve performance (Config F).

We also compare with other target domain augmentation methods, including Style Transfer and ASM. Specifically, these are employed to enlarge the target domain examples and then we constrained fine-tune our model using the augmented data. As indicated in Table~\ref{tab:oneshot-compare},
our method significantly facilitates the model adapting to the target domain, thus achieve better super-resolution performance.

\input{tables/ablations}
\input{tables/oneshot-compare}

%% file: tables/ablations.tex
\begin{table}[t]
\centering
\caption{Ablations on different configurations of the network architecture (A,B,C) and different configurations of the adaptation algorithm (D,E,F). $\uparrow$ indicates the higher the better, and $\downarrow$ indicates the lower the better.}\label{tab:ablations}
\footnotesize{
\setlength\tabcolsep{4pt}
\renewcommand\arraystretch{1}
\begin{tabular}{|l@{\hspace{1.5mm}}l|cccc|}
\thickhline
\rowcolor{LightGray}
& & \multicolumn{4}{c|}{\textbf{CelebA $\rightarrow$ ExtendedYaleB}} \\
\rowcolor{LightGray}
& \multirow{-2}{*}{\textbf{Configuration}} & LPIPS $\downarrow$ & FIQ $\uparrow$ & PSNR $\uparrow$ & SSIM $\uparrow$ \\ 
\hline
 & Baseline network & $0.48$ & $0.28$ & $17.64$ & $0.44$ \\
\hline
A & + Multi-scale features & $0.46$ & $0.30$ & $17.82$ & $0.44$ \\
B & + Multi-STN modules & $0.43$ & $0.34$ & $18.41$ & $0.45$ \\
C & + Predict offset scale & $0.38$ & $0.41$ & $20.39$ & $0.49$ \\
\hline
D & + Style mixing examples & $0.38$ & $0.41$ & $21.97$ & $0.52$ \\
E & + Soft mixing weight & $0.38$ & $0.42$ & $22.10$ & $0.54$ \\
F & + Constrained adaptation & $0.36$ & $0.46$ & $22.32$ & $0.55$ \\
\thickhline
\end{tabular}
}
\vspace{-2mm}
\end{table}

%% file: tables/oneshot-compare.tex
\begin{table}[t]
\centering
\caption{Comparisons on one-shot adaptation augmentation strategies. $\uparrow$ indicates the higher the better, and $\downarrow$ the lower the better.}\label{tab:oneshot-compare}
\footnotesize{
\setlength\tabcolsep{7pt}
\renewcommand\arraystretch{1}
\begin{tabular}{|l|cccc|}
\thickhline
\rowcolor{LightGray}
& \multicolumn{4}{c|}{\textbf{CelebA $\rightarrow$ ExtendedYaleB}} \\
\rowcolor{LightGray}
\multirow{-2}{*}{\textbf{Methods}} & LPIPS $\downarrow$ & FIQ $\uparrow$ & PSNR $\uparrow$ & SSIM $\uparrow$ \\ 
\hline
Direct fine-tuning & $0.44$ & $0.30$ & $20.11$ & $0.45$ \\
Style Transfer~\cite{huang2017arbitrary} & $0.42$ & $0.37$ & $20.16$ & $0.46$ \\
ASM~\cite{luo2020adversarial} & $0.40$ & $0.38$ & $20.71$ & $0.50$ \\
\hline
DAP-FSR (Ours) & $0.36$ & $0.46$ & $22.32$ & $0.55$ \\
\thickhline
\end{tabular}
}
\vspace{-4mm}
\end{table}